# Automatic speech recognition for launch control center communication using recurrent neural networks with data augmentation and custom language model

Kyongsik Yun*, Joseph Osborne, Madison Lee, Thomas Lu, Edward Chow
Jet Propulsion Laboratory, California Institute of Technology
4800 Oak Grove Drive, Pasadena, CA 91109

## ABSTRACT

Transcribing voice communications in NASA's launch control center is important for information utilization. However, automatic speech recognition in this environment is particularly challenging due to the lack of training data, unfamiliar words in acronyms, multiple different speakers and accents, and conversational characteristics of speaking. We used bidirectional deep recurrent neural networks to train and test speech recognition performance. We showed that data augmentation and custom language models can improve speech recognition accuracy. Transcribing communications from the launch control center will help the machine analyze information and accelerate knowledge generation.

## 1. INTRODUCTION

NASA's launch control center at the Kennedy Space Center has played an essential role in NASA's human space flight program for nearly 50 years. All data collected from the hanger, assembly facility, and launch pad is sent to the launch control center. Because of the nature of sending humans to space, engineers working at the launch control center must make sure everything is working perfectly before launch. In addition to a complete knowledge of the system, engineers need the commitment and conviction of astronaut safety. Decision making by each engineer is important for a successful launch. Supporting technologies for engineers in this kind of high-risk situation will be useful. We believe that automatic speech recognition is one of the supporting technologies that can be used to accurately analyze the situation for engineers.

NASA also believes the importance of communication accuracy. The launch control room was renovated in 2013 to reduce noise and improve communication accuracy. Sound-absorbing ceiling tiles, acoustic wall covering, and new carpet were installed [1]. Automatic speech recognition is an effective way to increase communication accuracy and reduce the possibility of errors [2,3].

Automatic speech recognition has three main applications, including input/output devices, communication aids, and information retrieval. The performance of automatic speech recognition has been greatly improved by using deep neural networks mainly for input / output devices [3]. Industry leaders such as Amazon, Apple, Baidu, Google, Microsoft, and IBM are evolving to help the public easily use their products through automatic speech recognition. Conversational interfaces based on Amazon's Alexa and Google's Home have been enhanced to naturally acquire information, access web services, and issue commands [4]. A recent study showed that deep learning based automatic speech recognition achieved human level performance in conversational speech recognition [5]. However, the accuracy of speech recognition in special purpose communication in noisy environments is not optimal [6].

The language that engineers use in the launch control center is a unique language for shuttle processing. They use many special terms and abbreviations that are difficult to decipher without prior knowledge of communication. In such an environment, speech recognition is important in terms of reducing the risk of misunderstanding, disseminating knowledge, and improving the launch protocol. Therefore, general-purpose speech recognition may not work. To improve the accuracy of automatic speech recognition in the verbal communication of professionals, we need to consider a customized language model that can describe the relationship between words and abbreviations in sentences [7].

We believe that the combination of training data, deep neural networks, data augmentation, and a customized language model is the key to improving automatic speech recognition in the verbal communication of professionals.

*kyun@jpl.nasa.gov; phone 1 818 354-1468; fax 1 818 393-6752; jpl.nasa.gov

## 2. TRAINING DATA AUGMENTATION

The recent development of deep learning-based speech recognition showed that conventional phoneme-based complex speech recognition is not required [8]. Rather, the system can output the final transcript directly based on the graphemes (i.e., alphabets) [9]. Previous studies used open-source data sets that typically contained thousands of hours of voice data [10]. State-of-the-art commercial speech recognition utilized 100,000 hours of data for training [11]. However, we have particularly small data sets because of the high cost of manual transcription (creation of ground truths) in specialized languages.

We trained our bidirectional recurrent neural networks using 575 unique sentences from NASA's mission control center. Nine lab members then created the voice data manually by saying the same sentence. We finally created total 2,000 speeches to analyze. Data augmentation techniques, including Gaussian noise (40 ~ 50dB), speech speed (0.9-1.1X), and volume modulation (-10 ~ 10dB) were applied to obtain augmented data corresponding to 10 times of original data [11,12]. Of the augmented data, the data including Gaussian noise was 60% and the speed and volume modulation data was 20% each.

## 3. BIDIRECTIONAL RECURRENT NEURAL NETWORKS

We used bidirectional recurrent neural networks (RNN) to train and test speech recognition performance. RNN can be thought of as an enhanced version of the hidden markov model (HMM) [13]. The HMM has a major drawback: the state is updated only from one state to the next so that the network cannot learn long-term dependencies [14]. For context-sensitive language decoding, it is important to learn long-term relationships between words. The solution is a recurrent neural network. Long-term dependence can be learned through back propagation [15].

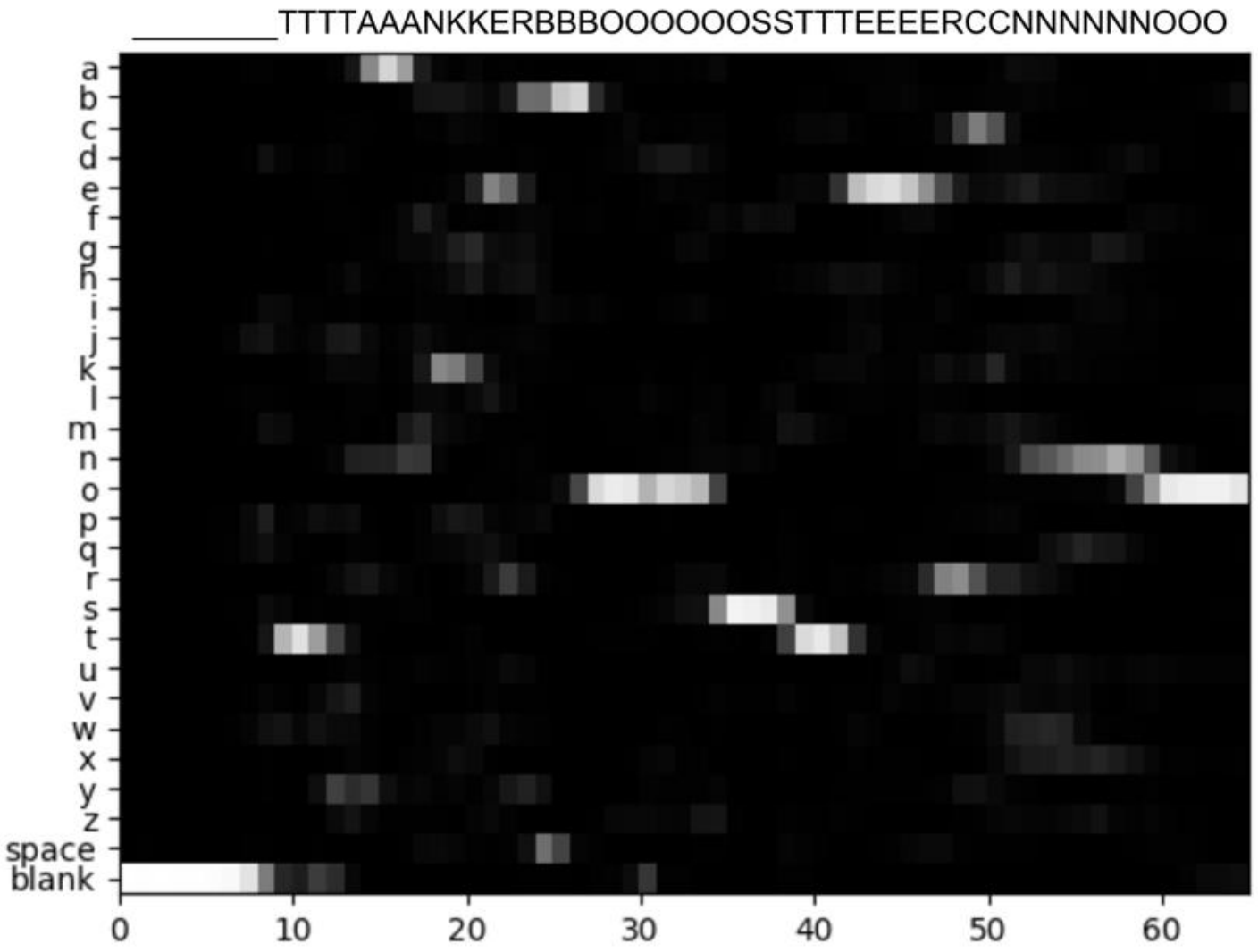

Figure 1. Recurrent neural network output. The X-axis represents the time step of each output of the recurrent neural module. The Y axis represents 26 alphabet characters (A-Z) and 2 special characters (blank and space), and consists of a total of 28 letters (graphemes). The intensity of the map represents the probability of the corresponding grapheme at each time step (white = 1, black = 0). The output is displayed at the top of the figure ("___ TTTAAANKKERBBBOOOSSTTEER ...") and further post processed to create the final transcript "TANKER BOOSTER".

Bidirectional RNNs collect information both in the past and in the future [16]. The original RNN, which obtains information entirely in the past, is not sufficient to accurately predict the current word, especially in situations where context is important. Given the sequence of words x(1), x(2), x(3), and x(4), the forward recurrent components can be represented by $\vec{a}(1)$, $\vec{a}(2)$, $\vec{a}(3)$, and $\vec{a}(4)$. x(1) is input to the forward recurrent component $\vec{a}(1)$. Output estimate is y(1).

Then the backward layer of the recurrent components are $\overleftarrow{a}(4)$, $\overleftarrow{a}(3)$, $\overleftarrow{a}(2)$, and $\overleftarrow{a}(1)$. x(1) is also input to the backward recurrent component $\overleftarrow{a}(1)$ outputting y(1). The forward recurrent networks compute from $\vec{a}(1)$ to $\vec{a}(4)$ direction, and the backward recurrent networks compute from $\overleftarrow{a}(4)$ to $\overleftarrow{a}(1)$. The predicted output y(1) is the result of combined network components of $\vec{a}(1)$ and $\overleftarrow{a}(1)$. The entire network consists of acyclic graphs. For example, the output y (3) is based on the past x(1) and x(2), the current x(3), and the future x(4). We used 3 convolutional input layers, 6 recurrent layers, 1 fully connected layer, and 1 softmax layer [11,12].

We used long short-term memory (LSTM) as a recurrent network component [17]. Conventional RNNs still have significant practical problems caused by the exponential decay of gradient descent that prevents learning long-term relationships between words. LSTM is a special type of recurrent neural network that can learn long-term dependencies through selective memory consolidation [18].

Then we used the connection temporal classification (CTC) cost for speech recognition output [19]. With RNN, the input sequence matches one-to-one with the RNN output. Therefore, the number of outputs is large and redundant. In speech recognition, the number of input information is generally much larger than the number of output characters. For example, five seconds of a 16,000 Hz audio input is 80,000 inputs, whereas the number of output characters per 5 seconds is at least 1,000 times less than 80,000 characters. Therefore, the key is to reduce redundancy. The basic rule of CTC cost function is to erase repeated characters that are not separated by a "blank". With CTC, the recurrent neural network output “___TTTAAANKKERBBBOOOSSTTEER” becomes “TANKERBOSTER” (Figure 1).

## 4. LANGUAGE MODEL

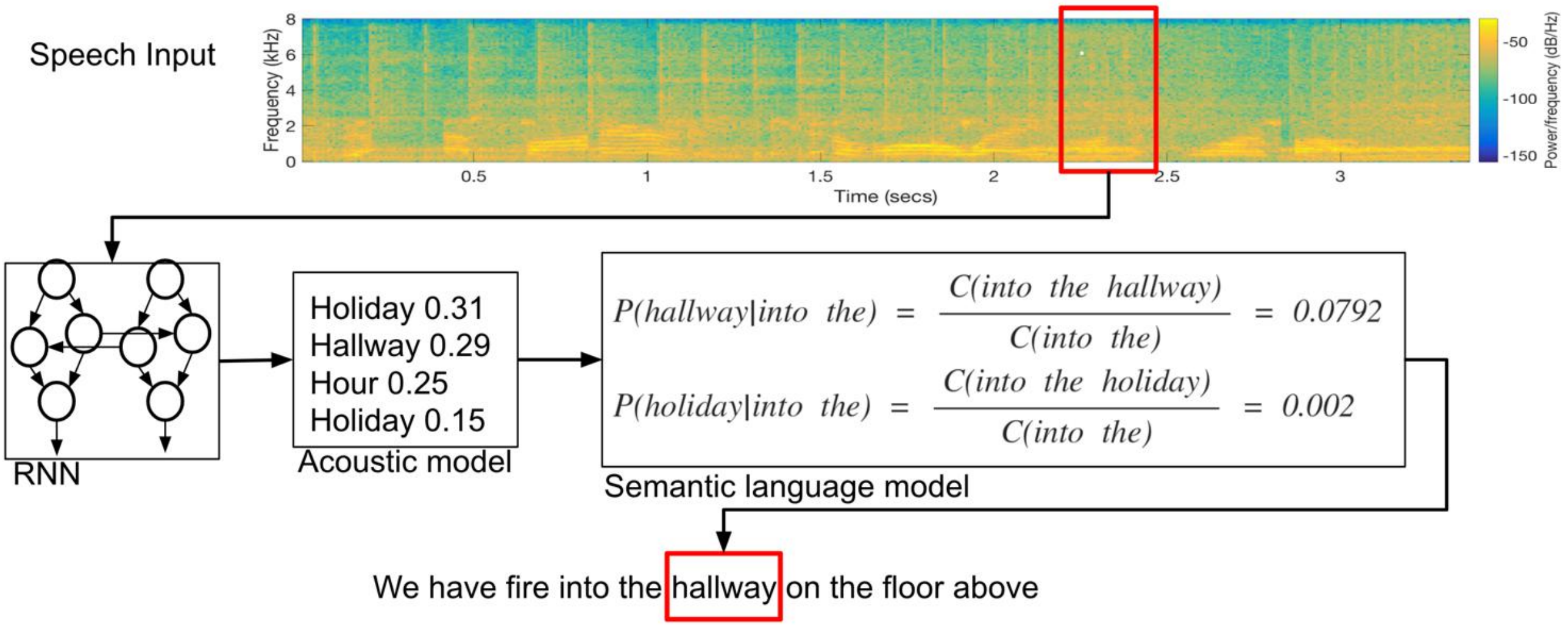

Figure 2. Speech recognition process. The audio input is converted to a time-frequency map. Each time window of the map is then entered into bidirectional recurrent neural networks (RNNs). Output candidates of the RNNs are ranked in the connectionist temporal classification (CTC). Candidates of the word are further processed and finally determined using the semantic language model.

We customized the language model to identify task-related abbreviations and technical terms (Figure 2). The language model uses the CTC output as input to return the probability of the last word in the given context of the previous words. The beam search algorithm was used to find the most accurate word candidates for speech recognition post-processing [20]. The algorithm considers three possible choices for the current word (beam width = 3). Unlike exact search algorithms, such as breadth first search or depth first search, beam search runs faster, but it is not guaranteed to find the exact maximum for $\arg\max_{y} P(y|x)$. Given the audio input *x*, *y(t)* is the output word at time *t*. Then the objective function is:

$$\arg\max_{y} \sum_{t=1}^{T_y} \log P(y(t)|x, y(1), \dots, y(t-1)) \quad (1)$$

The log was introduced to handle very small values of the probability, $P(y(t)|x, y(1), \dots, y(t-1))$, especially when the decoding sentence is long.

## 5. SPEECH RECOGNITION RESULTS IN NASA LAUNCH CONTROL CENTER DATA

We compared the word error rate (%) between Kaldi [21], CMUSphinx [22], IBM Watson Speech to Text, Google Speech API, RNN, and RNN with data augmentation and custom language models. Kaldi and CMUSphinx based on the hidden Markov models achieved word error rates of 53% and 39%, respectively. Commercial applications IBM Watson Speech to Text and Google Speech API (web-based solutions) achieved 45% and 31%, respectively. RNN alone had a word error rate of 48%. The combination of RNN and data augmentation achieved a word error rate of 26%. When combined RNN, data augmentation, and custom language model, the word error rate was the lowest at 18% (Figure 3).

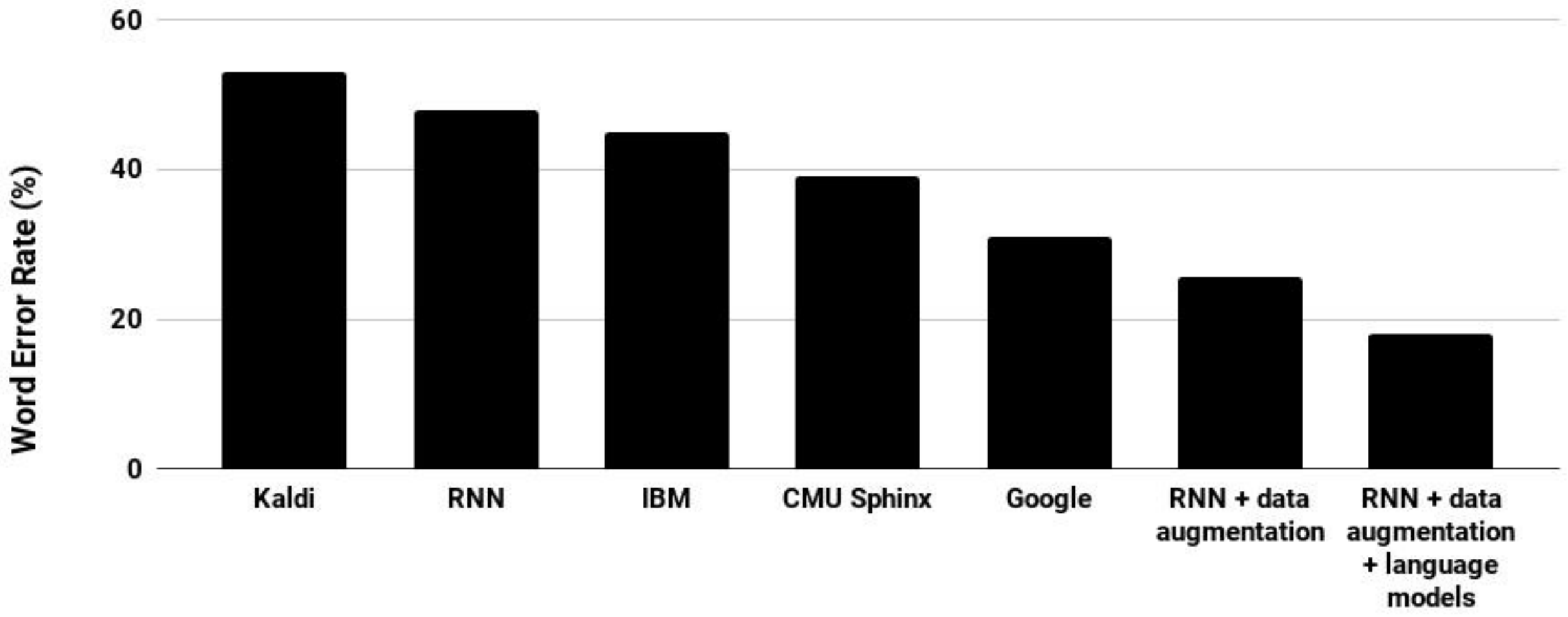

Figure 3. Speech recognition word error rate (%) of NASA's launch control center data. Open-source and commercial solutions did not perform well because of their highly specialized abbreviations and word orders. Only customized language models and data augmentation could improve accuracy.

## 6. HUMAN REVISION INTERFACE

We built a human revision interface to continually improve accuracy. NASA's Launch Control Center speech recording server transmits the speech stream to the client through the Session Description Protocol (SDP), a format that describes the streaming media communication parameters. We simultaneously analyze speech and speaker recognition through client-side analysis and then save the results in comma-separated values (CSV) data format. If necessary, we can access and modify the CSV dataset. After one modifies the dataset, the CSV data file is updated and the data is then sent to the server. The CSV data file is also used to improve the language model (Figure 4).

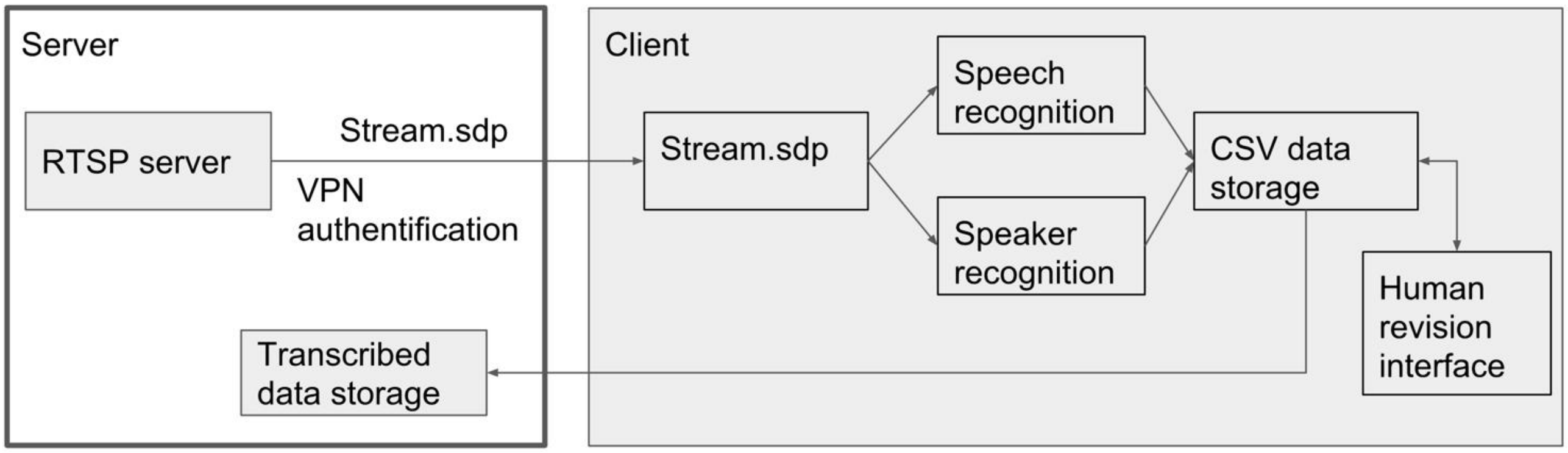

Figure 4. Speech / Speaker recognition with transcript revision interface. The Real-Time Streaming Protocol (RTSP) server records and streams speech data in real time through the Session Description Protocol (SDP). Access to the launch control center speech stream requires virtual private network (VPN) authentication. The client receives the SDP packet and then analyzes it to identify the speech and speakers. The transcribed data is then saved in the CSV (comma-separated value) file format, which we can manually modify to update the server's recorded data files and language model.

## 7. DISCUSSION

We found that bidirectional recurrent neural networks with data augmentation and custom language models achieved the best performance in highly specialized language environment with specialized abbreviation and grammar structures. The result is significant in that even perfect human hearing perception cannot fully comprehend communication at the launch control center without prior knowledge. A vast amount of training and knowledge of specific situations is important for transcription of professional voice communications.

The speech recognition problem is to map the audio input x to the script y. The human ear converts a one-dimensional audio input to the intensity of the frequency component. This can be thought of as a preprocessing step to generate a spectrogram that maps 2D information of the time and frequency of the audio input. The human brain utilizes a variety of contextual information to fully understand and perceive speech, including attention [23] and social cues [24,25], as well as sound itself.

By using attention models in the future, we can bias the state of recurrent neural networks to improve context-based speech recognition by paying more attention to specific words and phrases [26]. The attention model must be trained as a separate RNN in the previous state to determine how much attention should be paid to adjacent inputs to bias the current state of the primary RNN. Naturally, the processing time to train the network will be at least doubled. Because of its inefficiency, the attention model should be carefully considered for speech recognition solutions. The attention model can also be used for machine translation [27] and image caption with visual attention [28].

Automatic speech recognition, supported by computer vision, can be useful for improving speech recognition accuracy in noisy environments. Previous studies have shown that lip reading computer vision is far superior to traditional noise reduction methods [29-31]. Another possibility that can improve accuracy is to use background information such as location information [32] and history information [33]. This information has been shown to improve speech recognition accuracy by reducing possible word combinations in certain situations

Transcription by automatic speech recognition system and successive voice selection and correction by human have been proposed [34]. Adaptive and continuous improvement by humans in the loop is at the heart of an accurate speech recognition solution [35]. Human intervention can be activated only when low-confidence automatic transcription occurs [36]. In this study, we showed that human revision interfaces will help to continually improve speech recognition accuracy.

Automatic speech recognition is essential in natural human-robot interaction [37]. Automatic speech recognition can reduce the operational burden of professionals controlling robots in complex tasks, such as pilots and astronauts [38]. Especially in human space flight situations, the ease of communication is more important because of the restricted gravity and limited movement environment [39]. Moreover, humanoid robots with high accuracy of speech recognition through natural communication will help mitigate the potential psychological and psychiatric problems of astronauts in long-range human space exploration [40,41].

## 8. CONCLUSION

Data augmentation and customized language models helped improve speech recognition accuracy in challenging environments (e.g., limited training data size, highly specialized communication style, and word choice in the NASA launch control center). Using more data in the future with the human revision interface will improve the accuracy of speech recognition. By transcribing human verbal communication, machines will be able to analyze information and increase knowledge utilization.

## ACKNOWLEDGMENT

The research was carried out at the Jet Propulsion Laboratory, California Institute of Technology, under a contract with the National Aeronautics and Space Administration. The research was funded by the U.S. Department of Homeland Security Science and Technology Directorate Next Generation First Responders Apex Program (DHS S&T NGFR) under NASA prime contract NAS7-03001, Task Plan Number 82-106095.